\documentclass[runningheads]{llncs}

\usepackage{graphicx}
\usepackage{mwe}
\usepackage[export]{adjustbox}
\usepackage[utf8]{inputenc}
\usepackage{tcolorbox}
\usepackage{minted}
\usepackage[ruled]{algorithm2e}
\usepackage{hyperref}
\usepackage{bbding} 

\usepackage{array}
\usepackage{ragged2e}
\usepackage{graphicx} 
\newcolumntype{P}[1]{>{\RaggedRight\arraybackslash}p{#1}}

\begin{document}
\title{From Optimal Policies to \textit{Individual Differences}: Rethinking Reinforcement Learning for Biology}

\titlerunning{Rethinking Reinforcement Learning for Biology}
%
\author{Patrick Govoni\inst{1}\orcidID{0000-0001-8813-1387} \and
Palina Bartashevich\inst{2,3}\Envelope\orcidID{0000-0002-5908-8196} \and
Clémence Bergerot\inst{1}\orcidID{0000-0001-7640-283X} \and
Valerii Chirkov\inst{1,3}\orcidID{0000-0003-3950-898X} \and
Valentin Lecheval\inst{1,3}\orcidID{0000-0001-6041-2718} \and
Pawel Romanczuk\inst{1,3,4}\orcidID{0000-0002-4733-998X}}
\authorrunning{P. Govoni et al.}
%
\institute{Institute for Theoretical Biology, Humboldt Universität zu Berlin, Berlin, Germany \and 
Faculty of Life Sciences, Thaer-Institute for Agricultural and Horticultural Sciences, Humboldt Universität zu Berlin, Berlin, Germany \and
Science of Intelligence, Cluster of Excellence, Berlin, Germany \and
Bernstein Center for Computational Neuroscience, Berlin, Germany \\ 
\Envelope Corresponding author: \email{palina.bartashevich@hu-berlin.de}}
\maketitle              
\begin{abstract}
Reinforcement learning (RL) is primarily known as a computational method for optimizing control tasks, but it is increasingly used to explain biological behavior.
While RL successfully captures key aspects of biology, a major gap remains: between-agent behavioral variability.
Consistent individual differences naturally permeate biological populations, yet RL models typically present only the single best individual or the population average.
Addressing this gap requires moving beyond current practices to generate behavioral diversity using biologically plausible mechanisms.
Here, we examine approaches from various subfields of RL and outline potential paths forward to close the gap between biology and simulation.

\end{abstract}

\section{Introduction}
Reinforcement learning (RL) provides a computational framework where agents learn to make decisions by interacting with their environment to maximize cumulative rewards~\cite{Frankenhuis2019}.
Recently, RL has gained prominence in behavioral ecology to identify optimal behavioral strategies from an experience-driven, mechanistic perspective~\cite{Frankenhuis2019}. 
While RL offers a powerful way to study emergent adaptive behavior in complex environments, it relies on biologically implausible assumptions that limit its usefulness for biological theory.

This limitation arises because many RL applications focus on engineering a single optimal strategy (or policy), often framing the task as a concrete, isolated problem~\cite{clary2018variability}. 
In these applications, high learning variability is viewed as a hurdle to reproducibility and is commonly treated as noise to be averaged out.
While this approach is essential for engineering and control, it contrasts sharply with biological systems, 
where variability across individuals is not incidental but has meaningful evolutionary and ecological consequences
\cite{dallBehaviouralEcologyPersonality2004,montiglioSocialNicheSpecialization2013}.

Biological agents exhibit substantial and consistent individual differences in behavior, also referred to as behavioral heterogeneity or personality~\cite{dallBehaviouralEcologyPersonality2004}. 
These differences persisting across lifespans or generations may be structured along fundamental tradeoffs~\cite{leNeuralCorrelatesIndividual2024,wolfWhyPersonalityDifferences2014},
or simply emerge from individual experience~\cite{Bierbach2017,kosterTabulaRasaAgents2025,govoniVisuospatialNavigationDistance2024}.
Behavioral diversity is an fundamental component of natural selection, which relies on variation at multiple biological scales~\cite{yamamichi2023ecoevolutionary,bivortDevelopmentalOriginsBehavioral2025}.

Reporting only average or ``best'' outcomes in RL obscures the distribution of behaviors that can emerge during learning, hindering the detection of naturally behavioral diversity. 
While the machine learning literature promotes achieving policy diversity using ``top-down'' optimizations that explicitly reward novelty or variation (e.g.,~\cite{eysenbachDiversityAllYou2018,huHeterogeneityMultiAgentReinforcement2025}), we argue this is not the only path to diversity. 
Instead we advocate for a ``bottom-up'' approach, where individual differences emerge naturally via biologically plausible mechanisms~\cite{boogertMeasuringUnderstandingIndividual2018}.

We position RL as a framework for understanding the mechanisms of behavior in biological agents.
Here, we introduce a research program with behavioral heterogeneity serving as a key illustrative case to discuss how multiple aspects and features of current RL frameworks can be leveraged for the study of adaptive behavior.

In particular, we propose broadening the scope of RL along three relevant dimensions to capture individual heterogeneity from the bottom-up: 
expanding the decision space via exploration, multiple objectives, and distributional representations (Section~\ref{sec:beyond-single}); 
moving from isolated decisions to sequential dynamics, such as path-dependent biases, multiple timescales, and open-ended processes (Section~\ref{sec:learning-dynamics}); 
and embedding individual behavior in the population, through population-based algorithms, multi-agent environments, and social structures (Section~\ref{sec:social}).
For each of these directions, we review current RL features and abilities and identify how they can be applied to biological systems.
Pursuing these directions will not only help behavioral scientists model natural diversity,
but may also benefit the computational methods themselves by improving explainability and control.
We note that these research directions are complementary, non-exclusive, and may or may not apply to specific biological systems and questions.

\begin{figure}[ht!]
	\centering
	\includegraphics[width=.85\linewidth,trim={0 0 0 0},clip]{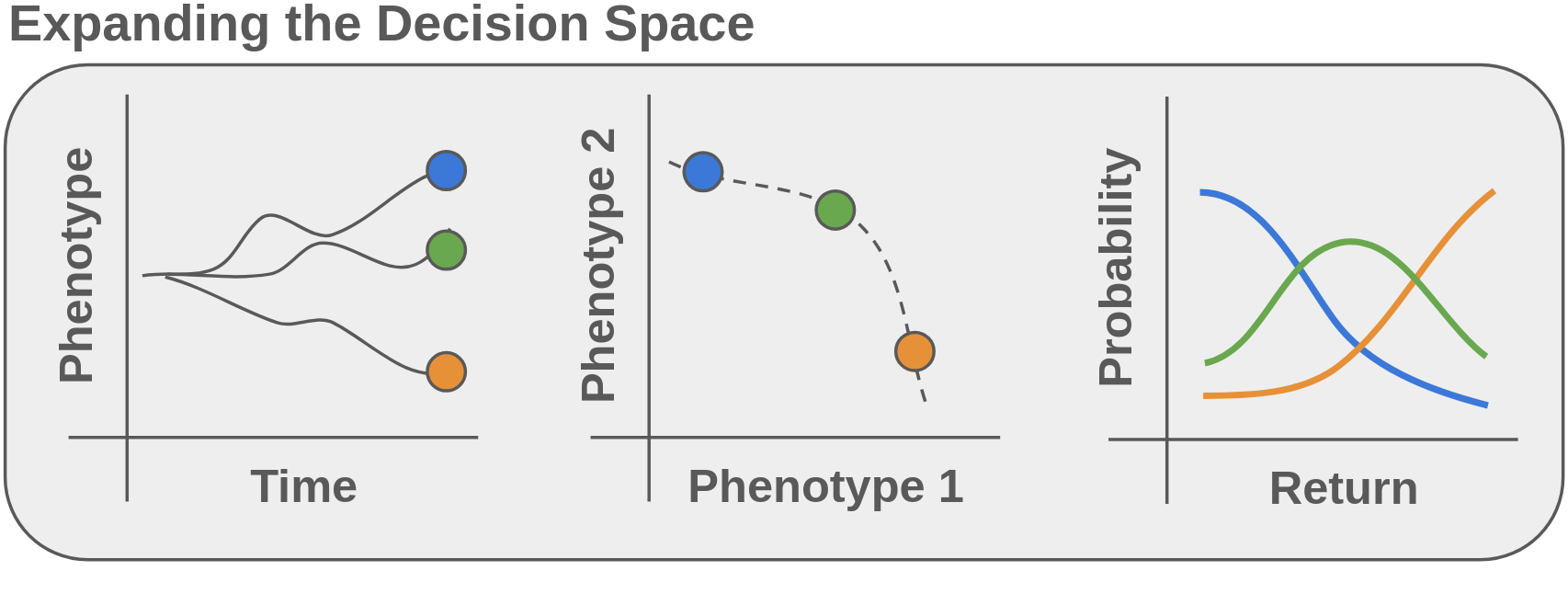}
	\includegraphics[width=.85\linewidth,trim={0 0 0 0},clip]{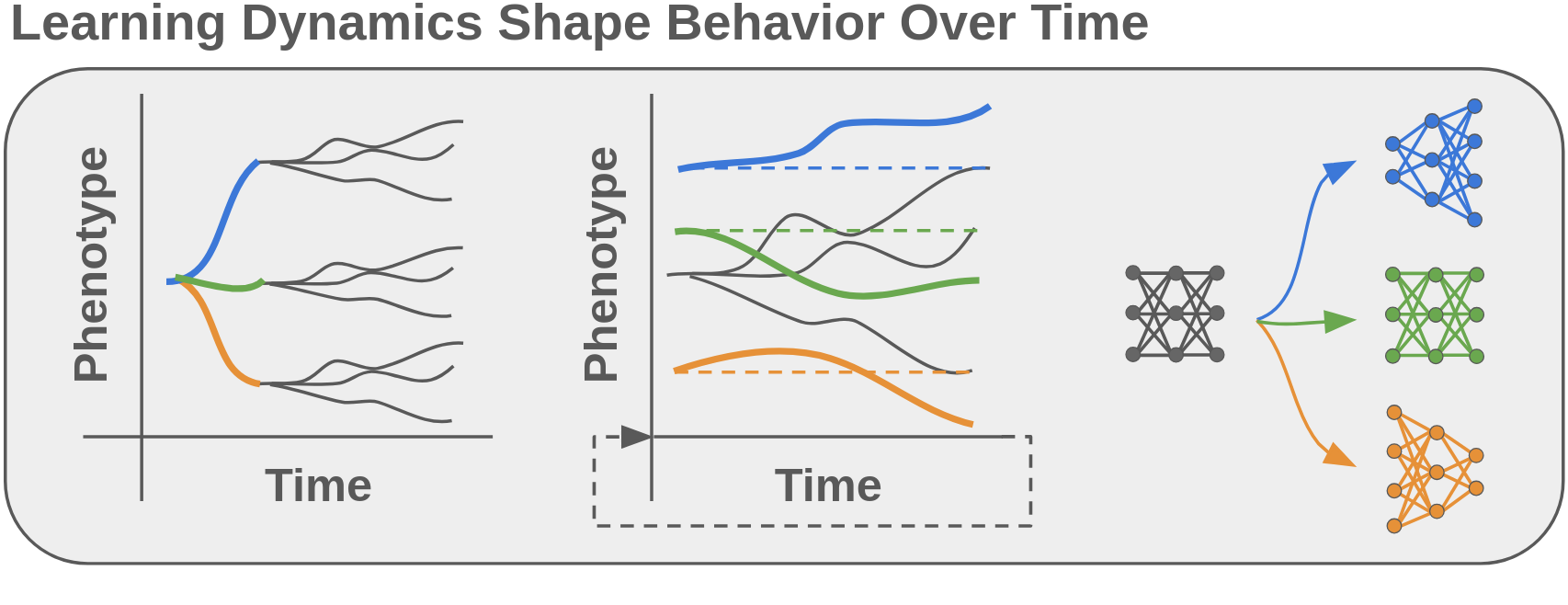}
	\includegraphics[width=.85\linewidth,trim={0 0 0 0},clip]{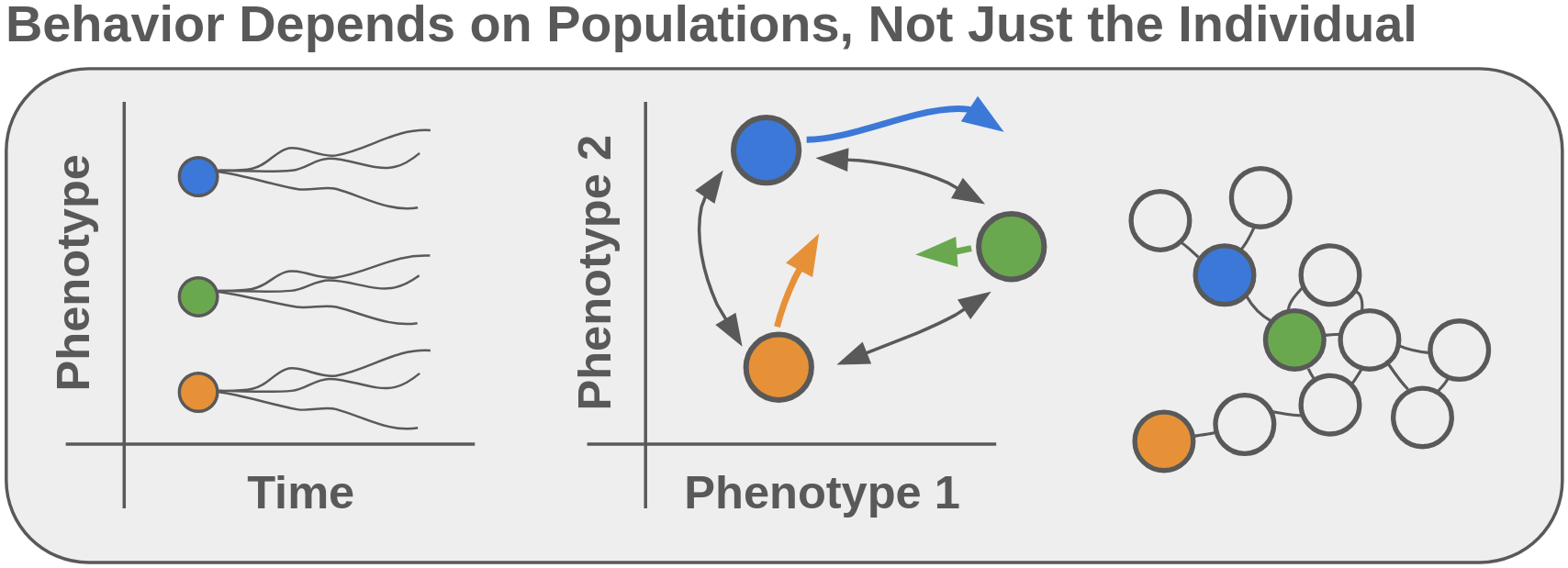}
	\caption{
        \textbf{Mechanisms for emergence of behavioral variation in RL.}
        Top: 
        (left) Agents may differentiate via simple random noise, environment interactions, or by optimizing non-reward dimensions; 
        (center) Pursuing multiple simultaneous objectives can segment unique individuals; 
        (right) Probability distributions of potential future return, or value, allow greater variability than scalar representations. 
        Middle:
        (left) Early perturbations push later dynamics into path-dependent, individually-biased behavior; 
        (center) Nesting multiple simulation timescales enables greater variance; 
        (right) Model state or action spaces that can change form open up behavioral dynamics. 
        Bottom:
        (left) Population-wide heterogeneity broadens potential behavioral differences; 
        (center) Interacting, decentralized multi-agent environments allow for social feedback to shape behavioral variation; 
        (right) Social structures, e.g. networks, provide an additional medium for differentiation.
		}
	\label{fig:main}
    \vspace{-7mm} 
  \end{figure}

\section{Expanding the Decision Space} 
\label{sec:beyond-single} 
\paragraph{\upshape\textbf{Exploration \& emergent behavior.} }
\label{subsec:exploration}
In RL, exploration addresses the problem of sparse rewards: 
agents pursuing exploitation alone tend to trap themselves in local optima,
making it necessary to vary behavior along dimensions not explicitly rewarded~\cite{Ladosz2022}.
Simple random action selection can generate the stochasticity needed for learning trajectories to diverge during training.
However, it is unclear whether randomness alone can consistently produce wide-ranging inter-individual differences as observed in biological systems.
More intrinsically motivated approaches to exploration use reward signals based on novelty, diversity, or entropy~\cite{pughQualityDiversityNew2016,eysenbachDiversityAllYou2018}, 
encouraging agents to deviate from their past experience. Exploration can also be defined relative to an agent’s internal model, driven by uncertainty minimization~\cite{mavrinDistributionalReinforcementLearning2019}. 
Taken together, the existence of multiple methods explicitly promoting behavioral variation
calls into question whether external reward-driven optimization alone is sufficient as a mechanism for the emergence of stable inter-individual differences.

Behavioral diversity can also emerge from agent-environment interactions. 
Research in ecological RL highlights how environmental structure shapes an agent's decisions~\cite{Co-reyes2020}, 
where initially identical agents may eventually develop different behavioral profiles through ongoing interactions with their surroundings. 
In this case, diversity is emergent rather than explicitly optimized. 
This reflects a biological principle whereby individual differences arise from the interactions between developmental dynamics and environmental constraints~\cite{bivortDevelopmentalOriginsBehavioral2025}.

\paragraph{\upshape\textbf{Multi-objective behavior.}} 
In general, biological behavior cannot be reduced to optimization with respect to a single objective.
Between-individual variation is often mapped along trade-offs between conflicting objectives, such as 
speed-accuracy~\cite{leNeuralCorrelatesIndividual2024}, cooperation-competition or social awareness~\cite{wolfWhyPersonalityDifferences2014}.
While task performance is often otherwise framed as finding a single optimal point, 
these problems typically feature a `Pareto front' of solutions
\cite{yangGeneralizedAlgorithmMultiObjective2019}, a set of equally optimal strategies where improving one goal (e.g., speed) necessarily requires sacrificing another (accuracy).
Since there is ``no free lunch" for modeling tradeoffs, scaling the relative importance of multiple goals becomes a choice, subjective to the decision-maker and subject to change with time or context. 

Multi-objective behavior may be implemented using multiple parallel models or a single conditional model~\cite{yangGeneralizedAlgorithmMultiObjective2019,abelsDynamicWeightsMultiObjective2019}. 
From a neuroscience perspective, individual networks trained on multiple tasks learn to project different tasks onto orthogonal manifolds~\cite{fleschOrthogonalRepresentationsRobust2022}, 
consistent with evidence of competition between different brain regions underlying individual differences along tradeoffs
\cite{leNeuralCorrelatesIndividual2024,wolfWhyPersonalityDifferences2014}.
The degree to which a network is able to maintain independent modules for different objectives or tasks depends on the computational capacity available and the potential to condense co-dependencies.
Energy constraints, such as parameter budgets, can provide a biologically plausible mechanism for forcing a choice between parallel models
\cite{bullmoreEconomyBrainNetwork2012}. While it is plausible that many aspects of reality can be modeled as single-objective reward schemes, a choice relating conflicting goals can allow sufficient complexity to uniquely specify an individual~\cite{vanchurinTheoryEvolutionMultilevel2022}.

\paragraph{\upshape\textbf{Distributional representation of value.}} 
In RL, ``value'' refers to the expected cumulative future reward associated with a state or action. 
However, typically used scalar-value representations discard information about uncertainty, variability, and risk by collapsing potentially rich distributions of future return into a single scalar quantity. 
For instance, two options with the same expected value may exhibit widely different variances, e.g., masking a choice between a low-risk, low-gain option and a high-risk, high-gain one. 

Distributional RL, by contrast, captures this difference by allowing agents to compute full probability distributions of outcomes~\cite{pmlr-v70-bellemare17a},
expanding the representational space not only from potential reward to uncertainty or risk, but also to temporal aspects such as delay~\cite{Sousa2025}. 
Similar to multi-objectivity, multi-dimensional representations permit a divergence in decision-making particular to an individual's experience and context
e.g., expanding potential choice to the optimistic~\cite{mavrinDistributionalReinforcementLearning2019,bergerotDeterministicDynamicsDistributional2025} 
or pessimistic~\cite{botvinickDepressionDisorderDistributional2025} sides of the distribution.
Furthermore, using a distributional representation of value has been found to depend on a specific sequence of rewards~\cite{tangNatureTemporalDifference2022},
suggesting a move toward a temporal, path-dependent understanding of individuality.

\section{Learning Dynamics Shape Behavior Over Time} 
\label{sec:learning-dynamics} 
\paragraph{\upshape\textbf{Path dependency and biases.}} 
While RL models are trained multiple times, varying random seeds and initializations, analysis typically reduces distinct learning trajectories to single-value snapshots, such as best or average performance. However, the strategy an agent uses changes across time, evolving across each training session~\cite{forestierIntrinsicallyMotivatedGoal2022}. Indeed, an agent interacts with its environment in a sequential, path-dependent manner, where unique individual experiences can push the learning process in a particular direction. 

A classic illustration is the ``hot stove effect''~\cite{denrell2001adaptation}, 
i.e., the tendency to neglect an option after an initial bad outcome. 
Touching a hot stove can give rise to a self-reinforcing loop: undersampling leads to incorrect valuation (``all stoves are hot''), driving further undersampling (``I will never come close to a stove again''). 
Similarly, a feedback loop can develop that over-prioritizes early interactions in a learning sequence~\cite{nikishinPrimacyBiasDeep2022},
or that confirms and perseveres with a belief that developed earlier in a training~\cite{bergerotModerateConfirmationBias2024}.
A bias can develop even with no explicit reward or punishment~\cite{kosterTabulaRasaAgents2025}.

Within the framework of dynamical systems, forming a bias can be represented as the tendency to fall into the basin of attraction of a previous choice, a local optimum,
even when other options are objectively more valuable~\cite{senftleben2021stay}. While these temporal asymmetries may either improve or worsen performance, 
convergence toward stable attractors through time may help to describe the formation of individual difference~\cite{bergerotModerateConfirmationBias2024}.

\paragraph{\upshape\textbf{Multi-timescale learning.}} 
Biological learning occurs over a range of entangled timescales: individual and evolutionary, short-term and across-lifetime, fast and slow decisions. RL commonly models behavior over a single timeline, ignoring potential recurrent effects. Positive feedback across time can amplify biases, as described above, thus learning processes coordinated across multiple timelines can compound difference, enabling greater behavioral divergence.

Meta-learning offers a formal framework for this idea, explicitly modeling decision-making in two nested temporal loops, where the inner (fast) loop acts across timesteps within a single task and the outer (slow) across task episodes~\cite{botvinickReinforcementLearningFast2019,nussenbaumUnderstandingDevelopmentReward2024}.
For example, a recent study separates an inner loop that directly optimizes rewards from an outer loop that does not receive direct reward feedback~\cite{Sandbrink2024CanRL}, 
linking to the intrinsically motivated exploration mechanisms described earlier~\cite{pughQualityDiversityNew2016,eysenbachDiversityAllYou2018}.

Separating learning across multiple timescales can help disentangle the effects of environmental context and developmental history~\cite{nussenbaumUnderstandingDevelopmentReward2024},
as well as unique cognitive contributions such as from goal-directed and habitual systems~\cite{Dezfouli2011}.
However these fast and slow decision processes, and their associated flexibility and inflexibility, may arise from a more fundamental separation in cognitive structure and objective~\cite{Han2023SynergizingHA}.

\paragraph{\upshape\textbf{Open-ended processes.}}
RL problem formulations are often restricted to a fixed state space, limiting agents to a predefined set of possible states~\cite{pessoaNetworksAdaptiveModels2026}.
Although learning temporal dynamics within a fixed state space is possible, behavior is limited by a predefined size and shape of its neural architecture as well as interactions.
In contrast, biological organisms adapt to unpredictable environments by continuously changing, growing, or condensing their internal model to fit the circumstances
\cite{nagabandiLearningAdaptDynamic2019}. 

Open-ended evolution provides an approach to such context-dependent adaptation, allowing models with dynamic state spaces and combinatorial flexibility
\cite{pessoaNetworksAdaptiveModels2026}. Differing from simply rewarding novelty or diversity~\cite{pughQualityDiversityNew2016,eysenbachDiversityAllYou2018}, open-endedness describes exploration at the structural level of model design. Iterating between testing and improving a model can help ground an accurate representation of reality and adapt it to shifting context
\cite{doncieuxOpenEndedLearningConceptual2018}.
Or rather, adaptivity can be achieved by continuously expanding a ``library'' of models, training a new model for each novel unexpected context~\cite{nagabandiDeepOnlineLearning2019}.

Navigating these boundless choices requires managing complexity. One approach is to temporarily restrict the decision space to a bounded stochastic action set
\cite{cohenDynamicPlanningOpenEnded2022}. To the contrary, an agent can learn to ``chunk'' useful action sequences to derive a broader set of possible options,
both temporally expanding potential behavioral choices and allowing an agent to nonlinearly stack experience
\cite{barretoOptionKeyboardCombining2019}.

\section{Behavior Depends On the Population, Not Just the Individual} 
\label{sec:social} 
\paragraph{\upshape\textbf{Population-based algorithms.}} 
While single-agent exploration has potential to induce behavioral variation, 
it is limited to learning from a single interaction history and reward gradient~\cite{nikishinPrimacyBiasDeep2022}. Population-based approaches bridge this barrier by utilizing the inherent heterogeneity in stepping outside the individual,
considering behavior as a distribution rather than a single policy
\cite{liBridgingEvolutionaryAlgorithms2024}.
Evolutionary or genetic algorithms do not explicitly follow a gradient, thus are more inclined to explore along neutral dimensions of variation, escaping potential local optima and favoring global convergence. 
Such exploration comes at the cost of lower efficiency and guarantees of convergence~\cite{liBridgingEvolutionaryAlgorithms2024}, 
but it is perhaps this ‘near-optimality’ that allows emergence of the diversity that we observe in natural systems~\cite{vanchurinTheoryEvolutionMultilevel2022}.

A few recent studies by our authors have found behavioral diversity to emerge from simple optimization using population-based algorithms
\cite{lechevalSmartSelfpropelledParticles2023,govoniVisuospatialNavigationDistance2024,govoniSocialspatialDependenciesLearning2026},
where a range of strategies was found to be equally performant. In one study~\cite{govoniVisuospatialNavigationDistance2024}, an intrinsic perceptual constraint (visual resolution) was identified as a key mechanism regulating variation between strategies. An extension of the model demonstrates how individuality further depends on social context, embodiment, and spatial localization~\cite{govoniSocialspatialDependenciesLearning2026}. It is currently untested whether variation remains when using an individual RL approach.

\paragraph{\upshape\textbf{Multi-agent environments.}} 
Embedding agents in multi-agent systems introduces non-stationary dynamics, where the environment changes as other agents act~\cite{marl-book}. While this complexity closely reflects biological environments, it violates the Markov property, a core assumption in traditional RL that environment dynamics remain stable and predictable, thus causing convergence issues. To address this, a common practice in multi-agent reinforcement learning (MARL) is to train agents centrally by sharing network parameters, 
inherently restricting behavioral individuality~\cite{wittIndependentLearningAll2020}.
While MARL research is primarily focused on coordinating agents to solve multi-agent tasks~\cite{marl-book}, 
its application to studying the emergence of biological individuality remains underexplored.

Some methods explicitly encourage differences among agents by maximizing or controlling heterogeneity across policies, observations, or objectives~\cite{liCelebratingDiversityShared2021,huHeterogeneityMultiAgentReinforcement2025}.
Like intrinsic rewards in single-agent RL, quantifying multi-agent diversity can bootstrap exploration and uncover complementary emergent strategies;
however, directly optimizing diversity in MARL is just as biologically implausible.

Emerging tools like graph neural networks enable fully decentralized learning, in which dynamic inter-agent communication develops alongside individual agent models
\cite{agarwalLearningTransferableCooperative2019,bettiniHeterogeneousMultiRobotReinforcement2023}. 
This communication, which may help agents naturally overcome non-stationarity, allows researchers to explicitly model both bottom-up heterogeneity in individual agent learning and emergent communication structures.
Ultimately, the key challenge for future work is to design MARL systems that can give rise to coordination and specialization in a decentralized manner, without relying on centralized training signals or explicitly engineered objectives.

\paragraph{\upshape\textbf{Influence of social structure.}} 
Multi-agent systems may produce behavioral diversity not only through the interaction of multiple agents, but also through specific interaction structures. From an evolutionary game theory perspective, choosing a consistent behavioral strategy supports one's predictability among interacting agents
\cite{dallBehaviouralEcologyPersonality2004,montiglioSocialNicheSpecialization2013,wolfWhyPersonalityDifferences2014}. 
Simply interacting more frequently may lead an agent to developing a preference, even if that preference is not explicitly rewarded
\cite{kosterTabulaRasaAgents2025}. Spatial as well as temporal ordering may stimulate divergence.
Asymmetries in the spatial layout of the environment can generate inequality in competitive settings~\cite{perolatMultiagentReinforcementLearning2017}.
Initializing agents close to each other, thus increasing perceptibility, can shift the balance of effective navigational strategy~\cite{govoniSocialspatialDependenciesLearning2026}.

Social network structures can favor behavioral variability when local connections dominate
\cite{huAchievingCoordinationMultiAgent2017}, or when interactions in multilayer systems are asymmetric, leading to the emergence of coalitions or hierarchies as conflicting objectives converge or diverge
\cite{huangAdaSocietyAdaptiveEnvironment2024,montiglioSocialNicheSpecialization2013}.
Rewarding an agent's position in the network itself can lead to asymmetric structural patterning, driving behavioral diversity \cite{chenEmergentCollectiveIntelligence2022}.

Extending these concepts to human institutions, much existing research assumes heterogeneity as an input to models. However, the existence of a shared medium such as the economy can lead to a diversity of social niches and behaviors \cite{perolatMultiagentReinforcementLearning2017,zhengAIEconomistTaxation2022}.
Political institutions can similarly stimulate varying degrees of diversity or conformity~\cite{pageWhereDiversityComes2014}.
However, RL-based models of governing bodies remain limited, although modeling cooperation over shared resources is a promising direction.

\begin{table}[t]
\caption{Synthesis of mechanisms linking RL concepts and behavioral individuality.}
\label{tab1}
\centering
\scalebox{0.9}{
\begin{tabular}{P{3.0cm} P{5.0cm} P{5.0cm}}
\hline
\multicolumn{1}{l}{\bfseries Mechanism} &
\multicolumn{1}{l}{\bfseries Biological interpretation} &
\multicolumn{1}{l}{\bfseries RL concept} \\
\hline\hline

Stochastic noise \& exploration &
Developmental noise, exploratory tendencies &
Random seeds, intrinsic motivation, agent--environment feedback \\

Multi-objective learning &
Behavioral trade-offs between \par conflicting goals &
Pareto optimization, reward weighting, energy constraints \\

Distributional value representation &
Risk-sensitive \& context-dependent valuation &
Return distributions (non-scalar), uncertainty/risk, temporal delay \\

\hline

Path-dependence &
Developmental history, \par experience-dependent biases &
Feedback within learning \par trajectories \\

Multi-timescale learning &
Development across nested timescales (evolution/ontogeny) &
Meta-learning, inner/outer nested loops \\

Open-ended representations &
Neural plasticity, adaptive internal models &
Dynamic state/action spaces \\

\hline

Population-based learning &
Population selection/variation &
Genetic/evolutionary algorithms \\

Multi-agent learning &
Social interaction, specialization, coordination/competition &
Diversity maximization, emergent roles, decentralized policies \\

Social structure &
Social organization/institutions, shared resources &
Interaction topology, \par communication networks \\

\hline
\end{tabular}
}
\vspace{-5mm}
\end{table}

\section{Discussion and Conclusion}
Consider the example of a single honeybee worker.
It must continuously adapt to a highly dynamic environment, balancing private and social cues to find which tasks best serve both itself and the colony within the current context.
As the worker ages and accumulates experience, tasks shift from cleaning and nursing within the nest, to foraging outside~\cite{johnson2010division},
where performance depends on its unique history, the specific sequence of cues it has encountered, and external pressures acting on its colony.
From the standpoint of the colony, its optimal distribution of tasks changes by the hour, both emerging from the heterogeneous behaviors across all workers and directing their continued dynamics.
While our paper outlines several distinct aspects of generating inter-individual variation in RL systems, biological systems integrate many, if not all, of these processes simultaneously.

The dimensions for generating individual differences explored in this paper are entangled. Rather than acting as independent sources of variability, they reflect different facets of a coupled learning–environment system.
Accordingly, the RL frameworks discussed can be applied in two ways: individually, to isolate and study specific mechanisms, or jointly, to capture interacting sources of behavioral variability when a single framework is not sufficient.

Computationally enabling the potential for individuality in RL might involve broadening the decision space along non-rewarded dimensions, between conflicting objectives, or among various representations of value. 
Examining temporal dynamics offers fresh perspectives on path-dependencies, multiple timescales, and open-ended processes. 
Or perhaps individuality may be more easily found at the population level, calling for individual learning algorithms, multi-agent environments, or structured social interactions. 
The number of possible equivalent solutions is immense. 
An individual may be determined by the Anna Karenina principle: every non-optimal behavioral strategy is unique in its own way~\cite{zaneveldStressStabilityApplying2017}.

Early behavioral ecologists treated inter-individual differences as non-adaptive noise relative to population-level average behavior, only later finding the individual level essential to understand behavioral evolution~\cite{dallBehaviouralEcologyPersonality2004,boogertMeasuringUnderstandingIndividual2018}. 
RL may serve to complement this pursuit with a framework that considers variability as a fundamental property of adaptive systems. Rather than viewing diversity as implicit stochasticity to be averaged out, or as an explicit objective to be optimized, we argue that behavioral individuality emerges from contingencies and feedback. Shaped by individual, environmental, and social dynamics, random variation can be amplified and reinforced over time into stable behavioral biases. Critical outstanding questions are when and how such transient divergence can stabilize into persistent behavioral individuality.

Currently, there is limited interaction between the communities developing RL as a computational tool and those studying behavioral development, as they tend to focus on different research questions: application of optimized behavior and understanding biological behavior. We argue that these perspectives are not mutually exclusive, and can instead form a feedback loop, where modeling and empirical observation iteratively inform and refine one another. 
This paper aims to act as a bridge and a catalyst for that collaboration.

We aim to provide guidance on navigating the RL landscape to study phenomena like inter-individual differences (Table~\ref{tab1}). Similar behavioral outcomes may arise from different underlying mechanisms, so RL as a tool should be used not only to reproduce behavior but also to probe various mechanistic explanations. Closer collaboration with computer scientists may ease the potential to configure plausible mechanisms to answer biological questions. Our hope is to move the field beyond using RL merely as a fitting tool, positioning it instead as a mechanistic framework for understanding the very origins of behavior. 

\vspace{0.3cm}
\noindent
\textbf{Acknowledgement.} 
\small This study was funded by the Deutsche Forschungsgemeinschaft (DFG, German Research Foundation) under Germany’s Excellence Strategy – EXC 2002/1 “Science of Intelligence” – project number 390523135. The funder played no role in study design, data collection, analysis and interpretation of data, or the writing of this manuscript. 

\bibliographystyle{unsrt}
\bibliography{literature}

@article{Frankenhuis2019,
title = {Enriching behavioral ecology with reinforcement learning methods},
journal = {Behavioural Processes},
volume = {161},
pages = {94-100},
year = {2019},
issn = {0376-6357},
url = {doi.org/10.1016/j.beproc.2018.01.008},
author = {Willem E. Frankenhuis and Karthik Panchanathan and Andrew G. Barto}
}

@article{lechevalSmartSelfpropelledParticles2023,
  title = {Smart Self-Propelled Particles: A Framework to Investigate the Cognitive Bases of Movement},
  shorttitle = {Smart Self-Propelled Particles},
  author = {Lecheval, Valentin and Mann, Richard P.},
  year = 2023,
  journal = {J R Soc Interface},
  volume = {20},
  number = {204},
  pages = {20230127},
  publisher = {Royal Society},
  url = {doi.org/10.1098/rsif.2023.0127}
}

@inproceedings{chenEmergentCollectiveIntelligence2022,
  title = {Emergent Collective Intelligence from Massive-Agent Cooperation and Competition},
  booktitle = {Deep RL Workshop NeurIPS},
  author = {Hanmo, Chen and et al.},
  year = 2022,
  url = {openreview.net/forum?id=KCWm-HV0PQT},
  langid = {english}
}

@article{Bierbach2017,
author = {Bierbach, David and Laskowski, Kate and Wolf, Max},
year = {2017},
pages = {15361},
title = {Behavioural individuality in clonal fish arises despite near-identical rearing conditions},
volume = {8},
journal = {Nat Commun},
url = {doi.org/10.1038/ncomms15361}
}

@inproceedings{clary2018variability,
  title = {Let's {{Play Again}}: {{Variability}} of {{Deep Reinforcement Learning Agents}} in {{Atari Environments}}},
  year = 2019,
  shorttitle = {Let's {{Play Again}}},
  booktitle = {{{NeurIPS 2018 Critiquing}} and {{Correcting Trends Workshop}}},
  author = {Clary, Kaleigh and Tosch, Emma and Foley, John and Jensen, David},
  url = {doi.org/10.48550/arXiv.1904.06312},
}

@inproceedings{huAchievingCoordinationMultiAgent2017,
  title = {Achieving {{Coordination}} in {{Multi-Agent Systems}} by {{Stable Local Conventions}} under {{Community Networks}}},
  booktitle = {Proc. 26th IJCAI},
  author = {Shuyue, Hu and Ho-fung, Leung},
  year = 2017,
  pages = {4731--4737},
  publisher = {AAAI Press},
  url = {doi.org/10.24963/ijcai.2017/659},
}

@article{vanchurinTheoryEvolutionMultilevel2022,
  title = {Toward a Theory of Evolution as Multilevel Learning},
  author = {Vanchurin, Vitaly and Wolf, Yuri I. and Katsnelson, Mikhail I. and Koonin, Eugene V.},
  year = 2022,
  journal = {Proc Natl Acad Sci},
  volume = {119},
  number = {6},
  pages = {e2120037119},
  issn = {0027-8424, 1091-6490},
  url = {doi.org/10.1073/pnas.2120037119},
}

@misc{Co-reyes2020,
title={Ecological Reinforcement Learning},
author={John D. Co-Reyes and Suvansh Sanjeev and Glen Berseth and Abhishek Gupta and Sergey Levine},
year={2020},
url={openreview.net/forum?id=S1xxx64YwH}
}

@inproceedings{Sandbrink2024CanRL,
  title={Can reinforcement learning model learning across development? Online lifelong learning through adaptive intrinsic motivation},
  author={Kai J Sandbrink and Brian Christian and Linas Nasvytis and Christian Schroeder de Witt and Patrick Butlin},
  booktitle={Proc. 36th CogSci},
  year={2024}
}

@article{zaneveldStressStabilityApplying2017,
  title = {Stress and Stability: Applying the {{Anna Karenina}} Principle to Animal Microbiomes},
  shorttitle = {Stress and Stability},
  author = {Zaneveld, Jesse R. and McMinds, Ryan and Vega Thurber, Rebecca},
  year = 2017,
  journal = {Nature Microbio},
  volume = {2},
  number = {9},
  pages = {17121},
  publisher = {Nature Publishing Group},
  issn = {2058-5276},
  url = {doi.org/10.1038/nmicrobiol.2017.121},
}

@article{nussenbaumUnderstandingDevelopmentReward2024,
  title = {Understanding the Development of Reward Learning through the Lens of Meta-Learning},
  author = {Nussenbaum, Kate and Hartley, Catherine A.},
  year = 2024,
  journal = {Nat Rev Psych},
  volume = {3},
  number = {6},
  pages = {424--438},
  publisher = {Nature Publishing Group},
  issn = {2731-0574},
  url = {doi.org/10.1038/s44159-024-00304-1},
}

@misc{bergerotDeterministicDynamicsDistributional2025,
  title = {Deterministic Dynamics of Distributional Multi-Agent Reinforcement Learning},
  author = {Bergerot, Cl{\'e}mence and Romanczuk, Pawel and Barfuss, Wolfram},
  year = 2025,
  primaryclass = {New Results},
  pages = {2025.12.22.696014},
  publisher = {bioRxiv},
  issn = {2692-8205},
  url = {doi.org/10.64898/2025.12.22.696014},
}

@article{Dezfouli2011,
    url = {doi.org/10.1371/journal.pcbi.1002055},
    author = {Keramati, Mehdi AND Dezfouli, Amir AND Piray, Payam},
    journal = {PLOS Comp Bio},
    publisher = {Public Library of Science},
    title = {Speed/Accuracy Trade-Off between the Habitual and the Goal-Directed Processes},
    year = {2011},
    volume = {7},
    pages = {1-21},
    number = {5},
}

@article{Han2023SynergizingHA,
  title={Synergizing habits and goals with variational Bayes},
  author={Dongqi, Han and Kenji Doya and Dongsheng, Li and Jun Tani},
  journal={Nat Commun},
  year={2023},
  volume={15},
  url={doi.org/10.1038/s41467-024-48577-7}
}

@article{boogertMeasuringUnderstandingIndividual2018,
  title = {Measuring and Understanding Individual Differences in Cognition},
  author = {Boogert, Neeltje J. and Madden, Joah R. and {Morand-Ferron}, Julie and Thornton, Alex},
  year = 2018,
  journal = {Philos Trans R Soc Lond B Biol Sci},
  volume = {373},
  number = {1756},
  pages = {20170280},
  issn = {0962-8436},
  url = {doi.org/10.1098/rstb.2017.0280},
}

@article{dallBehaviouralEcologyPersonality2004,
  title = {The Behavioural Ecology of Personality: Consistent Individual Differences from an Adaptive Perspective},
  shorttitle = {The Behavioural Ecology of Personality},
  author = {Dall, Sasha R. X. and Houston, Alasdair I. and McNamara, John M.},
  year = 2004,
  journal = {Ecology Letters},
  volume = {7},
  number = {8},
  pages = {734--739},
  issn = {1461-0248},
  url = {doi.org/10.1111/j.1461-0248.2004.00618.x},
}

@article{wolfWhyPersonalityDifferences2014,
  title = {Why Personality Differences Matter for Social Functioning and Social Structure},
  author = {Wolf, Max and Krause, Jens},
  year = 2014,
  journal = {Trends in Ecology \& Evolution},
  volume = {29},
  number = {6},
  pages = {306--308},
  publisher = {Elsevier},
  issn = {0169-5347},
  url = {doi.org/10.1016/j.tree.2014.03.008},
}

@article{leNeuralCorrelatesIndividual2024,
  title = {The {{Neural Correlates}} of {{Individual Differences}} in {{Reinforcement Learning}} during {{Pain Avoidance}} and {{Reward Seeking}}},
  author = {Thang, M. Le and Oba, Takeyuki and Couch, Luke and McInerney, Lauren and Chiang-Shan, R. Li},
  year = 2024,
  journal = {eNeuro},
  volume = {11},
  number = {2},
  publisher = {Society for Neuroscience},
  issn = {2373-2822},
  url = {doi.org/10.1523/ENEURO.0437-23.2024},
}

@misc{liBridgingEvolutionaryAlgorithms2024,
  title = {Bridging {{Evolutionary Algorithms}} and {{Reinforcement Learning}}: {{A Comprehensive Survey}} on {{Hybrid Algorithms}}},
  shorttitle = {Bridging {{Evolutionary Algorithms}} and {{Reinforcement Learning}}},
  author = {Pengyi, Li and et al.},
  year = 2024,
  number = {arXiv:2401.11963},
  eprint = {2401.11963},
  primaryclass = {cs},
  publisher = {arXiv},
  url = {doi.org/10.48550/arXiv.2401.11963},
}

@misc{huHeterogeneityMultiAgentReinforcement2025,
  title = {Heterogeneity in {{Multi-Agent Reinforcement Learning}}},
  author = {Hu, Tianyi and et al.},
  year = 2025,
  number = {arXiv:2512.22941},
  eprint = {2512.22941},
  primaryclass = {cs},
  publisher = {arXiv},
  url = {doi.org/10.48550/arXiv.2512.22941},
}

@misc{bettiniHeterogeneousMultiRobotReinforcement2023,
  title = {Heterogeneous {{Multi-Robot Reinforcement Learning}}},
  author = {Bettini, Matteo and Shankar, Ajay and Prorok, Amanda},
  year = 2023,
  number = {arXiv:2301.07137},
  eprint = {2301.07137},
  primaryclass = {cs},
  publisher = {arXiv},
  url = {doi.org/10.48550/arXiv.2301.07137},
}

@inproceedings{liCelebratingDiversityShared2021,
  title = {Celebrating {{Diversity}} in {{Shared Multi-Agent Reinforcement Learning}}},
  booktitle = {NeurIPS},
  author = {Chenghao, Li and et al.},
  year = 2021,
  volume = {34},
  pages = {3991--4002},
  publisher = {CAI},
}

@article{huangAdaSocietyAdaptiveEnvironment2024,
  title = {{{AdaSociety}}: {{An Adaptive Environment}} with {{Social Structures}} for {{Multi-Agent Decision-Making}}},
  shorttitle = {{{AdaSociety}}},
  author = {Yizhe, Huang and et al.},
  year = 2024,
  journal = {NeurIPS},
  volume = {37},
  pages = {35388--35413},
  url = {doi.org/10.52202/079017-1115},
}

@article{kosterTabulaRasaAgents2025,
  title = {Tabula Rasa Agents Display Emergent In-Group Behavior},
  author = {K{\"o}ster, Raphael and {Du{\'e}{\~n}ez-Guzm{\'a}n}, Edgar A. and Cunningham, William A. and Leibo, Joel Z.},
  year = 2025,
  journal = {Proc Natl Acad Sci},
  volume = {122},
  number = {25},
  url = {doi.org/10.1073/pnas.2319947121},
}

@inproceedings{perolatMultiagentReinforcementLearning2017,
  title = {A Multi-Agent Reinforcement Learning Model of Common-Pool Resource Appropriation},
  booktitle = {NeurIPS},
  author = {P{\'e}rolat, Julien and et al.},
  year = 2017,
  volume = {30},
  publisher = {CAI},
}

@article{zhengAIEconomistTaxation2022,
  title = {The {{AI Economist}}: {{Taxation}} Policy Design via Two-Level Deep Multiagent Reinforcement Learning},
  shorttitle = {The {{AI Economist}}},
  author = {Zheng, Stephan and Trott, Alexander and Srinivasa, Sunil and Parkes, David C. and Socher, Richard},
  year = 2022,
  journal = {Sci. Adv.},
  volume = {8},
  number = {18},
  publisher = {American Association for the Advancement of Science},
  url = {doi.org/10.1126/sciadv.abk2607},
}

@article{bivortDevelopmentalOriginsBehavioral2025,
  title = {The {{Developmental Origins}} of {{Behavioral Individuality}}},
  author = {de Bivort, Benjamin L.},
  year = 2025,
  journal = {Annual Review of Cell and Developmental Biology},
  volume = {41},
  number = {Volume 41, 2025},
  pages = {331--352},
  publisher = {Annual Reviews},
  issn = {1081-0706, 1530-8995},
  url = {doi.org/10.1146/annurev-cellbio-101323-025423},
}

@article{pageWhereDiversityComes2014,
  title = {Where Diversity Comes from and Why It Matters?},
  author = {Page, Scott E.},
  year = 2014,
  journal = {Eur. J. Soc. Psychol.},
  volume = {44},
  number = {4},
  pages = {267--279},
  issn = {1099-0992},
  url = {doi.org/10.1002/ejsp.2016},
}

@article{montiglioSocialNicheSpecialization2013,
  title = {Social Niche Specialization under Constraints: Personality, Social Interactions and Environmental Heterogeneity},
  shorttitle = {Social Niche Specialization under Constraints},
  author = {Montiglio, Pierre-Olivier and Ferrari, Caterina and R{\'e}ale, Denis},
  year = 2013,
  journal = {Philos Trans R Soc Lond B Biol Sci},
  volume = {368},
  number = {1618},
  pages = {20120343},
  issn = {0962-8436},
  url = {doi.org/10.1098/rstb.2012.0343},
}

@article{pughQualityDiversityNew2016,
  title = {Quality {{Diversity}}: {{A New Frontier}} for {{Evolutionary Computation}}},
  shorttitle = {Quality {{Diversity}}},
  author = {Pugh, Justin K. and Soros, Lisa B. and Stanley, Kenneth O.},
  year = 2016,
  journal = {Front. Robot. AI},
  volume = {3},
  publisher = {Frontiers},
  issn = {2296-9144},
  url = {doi.org/10.3389/frobt.2016.00040},
}

@misc{agarwalLearningTransferableCooperative2019,
  title = {Learning {{Transferable Cooperative Behavior}} in {{Multi-Agent Teams}}},
  author = {Agarwal, Akshat and Kumar, Sumit and Sycara, Katia},
  year = 2019,
  number = {arXiv:1906.01202},
  eprint = {1906.01202},
  primaryclass = {cs},
  publisher = {arXiv},
  url = {doi.org/10.48550/arXiv.1906.01202},
}

@article{bullmoreEconomyBrainNetwork2012,
  title = {The Economy of Brain Network Organization},
  author = {Bullmore, Ed and Sporns, Olaf},
  year = 2012,
  journal = {Nat Rev Neurosci},
  volume = {13},
  number = {5},
  pages = {336--349},
  publisher = {Nature Publishing Group},
  issn = {1471-0048},
  url = {doi.org/10.1038/nrn3214},
}

@article{forestierIntrinsicallyMotivatedGoal2022,
  title = {Intrinsically {{Motivated Goal Exploration Processes}} with {{Automatic Curriculum Learning}}},
  author = {Forestier, S{\'e}bastien and Portelas, R{\'e}my and Mollard, Yoan and Oudeyer, Pierre-Yves},
  year = 2022,
  journal = {Journal of Machine Learning Research},
  volume = {23},
  number = {152},
  pages = {1--41},
  issn = {1533-7928},
}

@misc{govoniVisuospatialNavigationDistance2024,
  title = {Visuospatial route-based navigation from the bottom-up},
  author = {Govoni, Patrick and Romanczuk, Pawel},
  year = 2026,
  url = {doi.org/10.48550/arXiv.2407.13535},
}

@misc{govoniSocialspatialDependenciesLearning2026,
  title = {Social-spatial dependencies for learning visual navigation},
  author = {Govoni, Patrick and Romanczuk, Pawel},
  year = 2026,
  url = {doi.org/10.48550/arXiv.2607.07460},
}

@inproceedings{yangGeneralizedAlgorithmMultiObjective2019,
  title = {A {{Generalized Algorithm}} for {{Multi-Objective Reinforcement Learning}} and {{Policy Adaptation}}},
  booktitle = {NeurIPS},
  author = {Runzhe, Yang and Xingyuan, Sun and Narasimhan, Karthik},
  year = 2019,
  volume = {32},
  publisher = {CAI},
}

@misc{eysenbachDiversityAllYou2018,
  title = {Diversity Is {{All You Need}}: {{Learning Skills}} without a {{Reward Function}}},
  shorttitle = {Diversity Is {{All You Need}}},
  author = {Eysenbach, Benjamin and Gupta, Abhishek and Ibarz, Julian and Levine, Sergey},
  year = 2018,
  number = {arXiv:1802.06070},
  eprint = {1802.06070},
  primaryclass = {cs},
  publisher = {arXiv},
  url = {doi.org/10.48550/arXiv.1802.06070},
}

@inproceedings{abelsDynamicWeightsMultiObjective2019,
  title = {Dynamic {{Weights}} in {{Multi-Objective Deep Reinforcement Learning}}},
  booktitle = {Proc. 36th ICML},
  author = {Abels, Axel and Roijers, Diederik and Lenaerts, Tom and Now{\'e}, Ann and Steckelmacher, Denis},
  year = 2019,
  pages = {11--20},
  publisher = {PMLR},
  issn = {2640-3498},
}

@article{fleschOrthogonalRepresentationsRobust2022,
  title = {Orthogonal Representations for Robust Context-Dependent Task Performance in Brains and Neural Networks},
  author = {Flesch, Timo and Juechems, Keno and Dumbalska, Tsvetomira and Saxe, Andrew and Summerfield, Christopher},
  year = 2022,
  journal = {Neuron},
  volume = {110},
  number = {7},
  pages = {1258-1270},
  publisher = {Elsevier},
  issn = {0896-6273},
  url = {doi.org/10.1016/j.neuron.2022.01.005},
}

@article{pessoaNetworksAdaptiveModels2026,
  title = {Beyond Networks: {{Toward}} Adaptive Models of Biological Complexity},
  shorttitle = {Beyond Networks},
  author = {Pessoa, Luiz},
  year = 2026,
  journal = {Physics of Life Reviews},
  volume = {56},
  pages = {67--81},
  issn = {1571-0645},
  url = {doi.org/10.1016/j.plrev.2025.11.007},
}

@misc{wittIndependentLearningAll2020,
  title = {Is {{Independent Learning All You Need}} in the {{StarCraft Multi-Agent Challenge}}?},
  author = {de Witt, Christian Schroeder and et al.},
  year = 2020,
  number = {arXiv:2011.09533},
  eprint = {2011.09533},
  primaryclass = {cs},
  publisher = {arXiv},
  url = {doi.org/10.48550/arXiv.2011.09533},
}

@inproceedings{barretoOptionKeyboardCombining2019,
  title = {The {{Option Keyboard}}: {{Combining Skills}} in {{Reinforcement Learning}}},
  shorttitle = {The {{Option Keyboard}}},
  booktitle = {NeurIPS},
  author = {Barreto, Andre and et al.},
  year = 2019,
  volume = {32},
  publisher = {CAI},
}

@misc{botvinickDepressionDisorderDistributional2025,
  title = {Depression as a Disorder of Distributional Coding},
  author = {Botvinick, Matthew and {Kurth-Nelson}, Zeb and Muller, Timothy and Dabney, Will},
  year = 2025,
  number = {arXiv:2507.16598},
  eprint = {2507.16598},
  primaryclass = {q-bio},
  publisher = {arXiv},
  url = {doi.org/10.48550/arXiv.2507.16598},
}

@article{tangNatureTemporalDifference2022,
  title = {The {{Nature}} of {{Temporal Difference Errors}} in {{Multi-step Distributional Reinforcement Learning}}},
  author = {Yunhao, Tang and et al.},
  year = 2022,
  journal = {NeurIPS},
  volume = {35},
  pages = {30265--30276},
}

@inproceedings{mavrinDistributionalReinforcementLearning2019,
  title = {Distributional {{Reinforcement Learning}} for {{Efficient Exploration}}},
  booktitle = {Proc. 36th ICML},
  author = {Mavrin, Borislav and Yao, Hengshuai and Kong, Linglong and Wu, Kaiwen and Yu, Yaoliang},
  year = 2019,
  pages = {4424--4434},
  publisher = {PMLR},
  issn = {2640-3498},
}

@InProceedings{pmlr-v70-bellemare17a,
  title = 	 {A Distributional Perspective on Reinforcement Learning},
  author =       {Marc G. Bellemare and Will Dabney and R{\'e}mi Munos},
  booktitle = 	 {Proc. 34th ICML},
  pages = 	 {449--458},
  year = 	 {2017},
  volume = 	 {70},
  publisher =    {PMLR}
}

@article{Sousa2025,
  title = {A multidimensional distributional map of future reward in dopamine neurons},
  volume = {642},
  ISSN = {1476-4687},
  url = {doi.org/10.1038/s41586-025-09089-6},
  number = {8068},
  journal = {Nature},
  publisher = {Springer Science and Business Media LLC},
  author = {Sousa,  Margarida and et al.},
  year = {2025},
  pages = {691–699}
}

@article{bergerotModerateConfirmationBias2024,
  title = {Moderate Confirmation Bias Enhances Decision-Making in Groups of Reinforcement-Learning Agents},
  author = {Bergerot, Cl{\'e}mence and Barfuss, Wolfram and Romanczuk, Pawel},
  year = 2024,
  journal = {PLOS Compu Bio},
  volume = {20},
  number = {9},
  publisher = {Public Library of Science},
  issn = {1553-7358},
  url = {doi.org/10.1371/journal.pcbi.1012404},
}

@article{botvinickReinforcementLearningFast2019,
  title = {Reinforcement {{Learning}}, {{Fast}} and {{Slow}}},
  author = {Botvinick, Matthew and Ritter, Sam and Wang, Jane X. and {Kurth-Nelson}, Zeb and Blundell, Charles and Hassabis, Demis},
  year = 2019,
  journal = {Trends in Cognitive Sciences},
  volume = {23},
  number = {5},
  pages = {408--422},
  publisher = {Elsevier},
  issn = {1364-6613, 1879-307X},
  url = {doi.org/10.1016/j.tics.2019.02.006},
}

@inproceedings{nikishinPrimacyBiasDeep2022,
  title = {The {{Primacy Bias}} in {{Deep Reinforcement Learning}}},
  booktitle = {Proc. 39th ICML},
  author = {Nikishin, Evgenii and Schwarzer, Max and D'Oro, Pierluca and Bacon, Pierre-Luc and Courville, Aaron},
  year = 2022,
  pages = {16828--16847},
  publisher = {PMLR},
  issn = {2640-3498},
}

@misc{nagabandiDeepOnlineLearning2019,
  title = {Deep {{Online Learning}} via {{Meta-Learning}}: {{Continual Adaptation}} for {{Model-Based RL}}},
  shorttitle = {Deep {{Online Learning}} via {{Meta-Learning}}},
  author = {Nagabandi, Anusha and Finn, Chelsea and Levine, Sergey},
  year = 2019,
  number = {arXiv:1812.07671},
  eprint = {1812.07671},
  primaryclass = {cs},
  publisher = {arXiv},
  url = {doi.org/10.48550/arXiv.1812.07671},
}

@misc{nagabandiLearningAdaptDynamic2019,
  title = {Learning to {{Adapt}} in {{Dynamic}}, {{Real-World Environments Through Meta-Reinforcement Learning}}},
  author = {Nagabandi, Anusha and et al.},
  year = 2019,
  number = {arXiv:1803.11347},
  eprint = {1803.11347},
  primaryclass = {cs},
  publisher = {arXiv},
  url = {doi.org/10.48550/arXiv.1803.11347},
}

@misc{cohenDynamicPlanningOpenEnded2022,
  title = {Dynamic {{Planning}} in {{Open-Ended Dialogue}} Using {{Reinforcement Learning}}},
  author = {Cohen, Deborah and et al.},
  year = 2022,
  eprint = {2208.02294},
  primaryclass = {cs},
  publisher = {arXiv},
  url = {doi.org/10.48550/arXiv.2208.02294},
}

@article{Ladosz2022,
author = {Ladosz, Pawel and Weng, Lilian and Kim, Minwoo and Oh, Hyondong},
title = {Exploration in deep reinforcement learning: A survey},
year = {2022},
publisher = {Elsevier Science Publishers B. V.},
volume = {85},
issn = {1566-2535},
url = {doi.org/10.1016/j.inffus.2022.03.003},
journal = {Inf. Fusion},
pages = {1–22},
numpages = {22}
}

@article{doncieuxOpenEndedLearningConceptual2018,
  title = {Open-{{Ended Learning}}: {{A Conceptual Framework Based}} on {{Representational Redescription}}},
  shorttitle = {Open-{{Ended Learning}}},
  author = {Doncieux, Stephane and et al.},
  year = 2018,
  journal = {Front. Neurorobot.},
  volume = {12},
  publisher = {Frontiers},
  issn = {1662-5218},
  url = {doi.org/10.3389/fnbot.2018.00059},
}

@article{senftleben2021stay,
  title={To stay or not to stay: The stability of choice perseveration in value-based decision making},
  author={Senftleben, Ulrike and Schoemann, Martin and Rudolf, Matthias and Scherbaum, Stefan},
  journal={Quarterly Journal of Experimental Psychology},
  volume={74},
  number={1},
  pages={199--217},
  year={2021},
  publisher={Sage Publications Sage UK: London, England}
}

@article{denrell2001adaptation,
  title={Adaptation as information restriction: The hot stove effect},
  author={Denrell, Jerker and March, James G},
  journal={Organization science},
  volume={12},
  number={5},
  pages={523--538},
  year={2001},
  publisher={INFORMS}
}

@article{johnson2010division,
  title={Division of labor in honeybees: form, function, and proximate mechanisms},
  author={Johnson, Brian R},
  journal={Behav Ecol Sociobiol},
  volume={64},
  number={3},
  pages={305--316},
  year={2010},
  publisher={Springer},
  url={doi.org/10.1007/s00265-009-0874-7}
}

@article{yamamichi2023ecoevolutionary,
author = {Yamamichi, Masato and Letten, Andrew D. and Schreiber, Sebastian J.},
title = {Eco-evolutionary maintenance of diversity in fluctuating environments},
journal = {Ecology Letters},
volume = {26},
year = 2023,
number = {S1},
pages = {S152-S167},
url = {doi.org/10.1111/ele.14286},
}

@book{marl-book,
  author = {Stefano V. Albrecht and Filippos Christianos and Lukas Sch\"afer},
  title = {Multi-Agent Reinforcement Learning: Foundations and Modern Approaches},
  publisher = {MIT Press},
  year = {2024}
}
\end{document}